# Real-time distortion prediction in metallic additive manufacturing via a physics-informed neural operator approach

Mingxuan Tian[a], Haochen Mu[a,b,c,*], Donghong Ding[a,b,c,*], Mengjiao Li[a], Yuhan Ding[a,b,c], Jianping Zhao[a]

[a] School of Mechanical and Power Engineering, Nanjing Tech University, Nanjing 211816, China
[b] Institute of Reliability centered Manufacturing, Nanjing Tech University, Nanjing 211816, China
[c] Jiangsu Provincial Key Laboratory of Energy Power Manufacturing Equipment and Reliability Technology, Nanjing Tech University, Nanjing 211816, China

*Corresponding author(s). E-mail(s): 202410007697@njtech.edu.cn, donghong@njtech.edu.cn;

**Abstract**

With the development of digital twins and smart manufacturing systems, there is an urgent need for real-time distortion field prediction to control defects in metal Additive Manufacturing (AM). However, numerical simulation methods suffer from high computational cost, long runtimes that prevent real-time use, while conventional Machine learning (ML) models struggle to extract spatiotemporal features for long-horizon prediction and fail to decouple thermo-mechanical fields. This paper proposes a Physics-informed Neural Operator (PINO) to predict z and y-direction distortion for the future 15 s. Our method, Physics-informed Deep Operator Network-Recurrent Neural Network (PIDeepONet-RNN) employs trunk and branch network to process temperature history and encode distortion fields, respectively, enabling decoupling of thermo-mechanical responses. By incorporating the heat conduction equation as a soft constraint, the model ensures physical consistency and suppresses unphysical artifacts, thereby establishing a more physically consistent mapping between the thermal history and distortion. This is important because such a basis function, grounded in physical laws, provides a robust and interpretable foundation for predictions. The proposed models are trained and tested using datasets generated from experimentally validated Finite Element Method (FEM). Evaluation shows that the model achieves high accuracy, low error accumulation, time efficiency. The max absolute errors in the z and y-directions are as low as 0.9733 mm and 0.2049 mm, respectively. The error distribution shows high errors in the molten pool but low gradient norms in the deposited and key areas. The performance of PINO surrogate model highlights its potential for real-time long-horizon physics field prediction in controlling defects.

## 1. Introduction

The digital transformation of manufacturing is increasingly driven by advances in Machine

Learning (ML) and Digital Twin (DT) technologies, which enable intelligent process monitoring, real-time decision-making, and predictive control[1]. As Industry 4.0 continues to evolve, metallic Additive Manufacturing (AM) has become a cornerstone of next-generation production, recognized for its ability to fabricate complex metal parts with exceptional design freedom and material efficiency[2]. Among metallic AM processes, Wire Arc Additive Manufacturing (WAAM), also referred to as Directed Energy Deposition-Arc (DED-Arc), has gained prominence due to its high deposition rate, scalability, and cost-effectiveness, making it suitable for large-scale structural applications in aerospace, maritime, energy, and heavy industry sectors[3, 4].

Despite these advantages, predicting and controlling distortion during the deposition process remains a challenge[5]. The highly concentrated heat input and rapid cooling in WAAM result non-uniform heat accumulation, which causes uneven thermal expansion and contraction within the material, thereby generating internal stresses[6]. When the stress exceeds the yield limit of the material, the component undergoes permanent plastic distortion, such as cracking or distortion[7]. Uncontrolled distortion not only compromises dimensional accuracy but also undermines structural integrity, potentially leading to failure in critical applications[8, 9]. The multi-physics interaction is inherently a coupled thermo-mechanical problem[10]. The spatiotemporal distribution of the temperature field is the primary cause of the initiation and evolution of distortion, while distortion also influence the temperature distribution[11]. Real-time distortion prediction involves understanding the complex relationship between thermo-mechanical history to predict long-horizon distortion distribution, which is crucial for distortion control and quality assurance in WAAM[12]. However, accurate real-time distortion prediction remains difficult due to the highly nonlinear and time-varying nature of thermo-mechanical coupling, as well as the temperature-dependent thermophysical properties.

Traditionally, researchers have relied on numerical simulation and data-driven methods to characterize the thermo-mechanical behavior of metallic AM. Numerical simulation methods usually employ the Finite Element Method (FEM)[13] and Computational Fluid Dynamics (CFD)[14]. FEM can capture detailed thermo-mechanical responses by establishing discretized numerical models of complex systems and solving the coupled governing equations. For example, Li et al.[15] developed a FEM model to investigate the thermal stress evolution and residual stress distribution in circular thin-walled parts. Unlike the FEM, CFD is focused on simulating the dynamic behavior of droplets and molten pools, integrating a variety of physical phenomena, and modeling multiscale phenomena spanning from the microscopic to the macroscopic levels. For instance, Chen et al.[16] developed three-dimensional CFD model to study the transient coupling behavior of heat and metal transfer, fluid flow, and solidified bead shape. It's extremely long computing time (e.g., 2 or 3 weeks for simulating 1 s of deposition[17, 18] ) hinders its applications in creating training datasets. Overall, numerical simulations lack adaptiveness as they need to be resolved when deposition scenario changes (e.g., changing deposition parameters or path, or calibrate with temperature feedback), which is the major limitation for their applications in real-time monitoring and adaptive control[19].

With the rapid advancement of artificial intelligence and ML, there has been a growing shift in metallic AM from numerical simulations toward data-driven surrogate model to overcome their high computational cost and limited adaptability[20, 21]. Data-driven ML models can learn potential nonlinear physical mappings from existing data, achieve real-time and flexible thermo-mechanical responses predictions for similar scenarios[22]. For example, Zhang et al.[23] achieved efficient prediction of residual stress in laser powder bed fusion by integrating a simplified FEM model with ML algorithms (e.g. multilayer perceptron, random forest). Hajializadeh et al.[24, 25] employed a modeling approach combining Artificial Neural Networks (ANN) with FEM to efficiently predict the residual stress distribution in AISI304L parts with different geometries fabricated by DED. Wu et al.[26] analyzed residual stress data from 243 alloy samples using random forest and neural networks, achieving 97% prediction accuracy and revealing the hierarchy of key variables affecting residual stress, with substrate preheating temperature identified as the most critical process parameter. To further extract the spatiotemporal features of the process data, Xie et al.[27] proposed a mechanistic data-driven framework that integrates wavelet transform and Convolutional Neural Network (CNN) to predict the location-dependent mechanical properties of manufactured parts based on thermal history. Zhou et al.[28, 29] developed an optimization method for continuous tool-path planning that enhances thermo-mechanical properties by employing a hybrid CNN and Recurrent Neural Network (RNN) architecture. Mu et al.[30] proposed an online distortion simulation model, which integrates a vector quantized variational autoencoder with RNN layers. However, these models typically achieve high predictive accuracy under similar deposition scenarios but exhibit limited capabilities in multi-physics decoupling, leading to low generalization across different deposition scenarios and error accumulation in long-horizon predictions.

Furthermore, historical distortion and temperature fields record irreversible processes such as heat accumulation and plastic strain evolution, whose cumulative effects directly determine the stress distribution and final distortion morphology in subsequent manufacturing stages[31]. Current distortion prediction methods for WAAM fail to adequately learn the evolution of distortion and temperature fields throughout the entire manufacturing history[25]. Most existing prediction models focus on utilizing process information (e.g., thermal cycles and process parameters) to estimate residual stresses and distortion, lacking the capability to model the real-time evolution of thermo-mechanical coupling[12]. Furthermore, current ML models are trained on fixed meshes, nodal representations, and strictly constrained boundary conditions, even minor variations in geometry, scanning strategy, or material properties can degrade their predictive accuracy[32]. Thus, decoupling thermal and mechanical histories is key to regulate data-driven models learning how distortion evolves in WAAM; and, a new path forward is to combine physical insight with Physics-informed Neural Operator (PINO) approach[33]. This study proposes an improved Physics-informed Deep Operator Network-RNN (PIDeepONet-RNN) model that incorporates PINO and RNN units, separates thermo-mechanical histories, and aims to accurately learn and predict the thermo-mechanical constitutive behavior of metal AM for the future time steps. The proposed surrogate model for real-time distortion prediction is expected to exhibit the following key features:

1. **Physics decoupled**: the developed surrogate model can achieve decoupling of the thermo-mechanical response in metal AM.
2. **Temporal analysis**: it can achieve history-informed learning while preserving the underlying physics of thermo-mechanical interactions.
3. **Long-horizon prediction**: the model should possess long-horizon prediction capability and exhibits low error accumulation over time.
4. **Model adaptiveness**: it can possess strong generalization capabilities to handle unknown process parameters and geometric evolution, while achieving high predicting accuracy

5. **Time-efficiency**: the decoupled architecture enables rapid convergence during the training phase, demonstrating better training and prediction efficiency to meet the requirements for real-time applications.

Physics-informed Machine Learning (PIML) embeds physical priors (e.g., governing equations, conservation laws) into the architecture or training process of ML models, effectively enhancing their generalization capability and prediction accuracy[34]. A typical implementation involves incorporating residual constraints from Partial Differential Equation (PDE) into the loss function, thereby ensuring that model outputs adhere to physical principles[35]. For example, Sharma et. al[36] proposed a thermodynamic Physics-informed Neural Network (PINN) framework, involves two separate networks for rapid prediction of thermal stress and temperature evolution during DED processes. However, such methods typically treat thermo-mechanical fields as coupled inputs and fail to explicitly decouple the complex interactions between thermal history and mechanical response, which limits their extrapolation capability and prediction efficiency in multi-physics coupled problems. Recently, operator learning methods represented by the Deep Operator Network (DeepONet) have gained significant attention due to their advantages in decoupling multi-physics fields.[37]. By processing inputs from different physical domains separately through its branch and trunk networks, DeepONet enables decoupled modeling of thermo-mechanical coupling problems. This decoupled architecture not only accelerates the training of PINN but also enhances their extrapolation and generalization capabilities. Researchers have explored the application of DeepONet in the field of metallic AM. For example, Jiao et al.[38] proposed a real-time welding temperature field prediction method that utilizes real-time welding current, voltage, and heat source parameters to update the heat source function, calculates the real-time heat flux density, and employs DeepONet model to solve the heat flux density for predicting temperature. However, the FNN-based DeepONet exhibits limitations in computational efficiency, flexibility, and capturing spatial correlations when handling complex geometries and high-dimensional inputs. Kushwaha et al.[39] proposed ResUNet-based DeepONet structure for multi-physics field prediction in AM. PINO incorporates physical conservation laws (e.g., PDE and boundary conditions) as prior knowledge, reducing reliance on large-scale labeled data and enhancing the model's stability and reliability in extrapolation and under varying boundary conditions[40], which is therefore used in this study.

The remainder of this paper is organized into four sections. Section 2 details the datasets for FEM modelling and the proposed model architectures. Section 3 outlines the experimental setup. Section 4 presents and evaluates the performance of the surrogate model. Finally, Section 5 summarizes the main contributions of this article and provides future outlook.

## 2. Methodology

This section describes the real-time distortion predictive surrogate modelling system, including datasets build, model architecture, and evaluation method.

2.1. System overview

The proposed real-time distortion prediction framework is shown in Figure 1, which organized into three modules: experimental data acquisition, FEM modelling, and surrogate model prediction. First, temperature and distortion histories are captured during WAAM process using thermal camera and laser scanning. Second, these experimental data are employed to calibrate parameters and results of the thermo-mechanical FEM model, such as geometric sizes and heat input. This process aligns the simulated results with physical measurements and ensures the reliability of both the FEM model and simulated physical fields. The calibrated FEM model was used to generate temperature and distortion fields for multiple process parameters; these outputs were collected and preprocessed to form the training and test datasets. Finally, the proposed surrogate model leverages a PINO approach jointly exploit thermal and distortion histories. This design captures long-horizon thermo-mechanical dependencies, enabling accurate, real-time long-horizon prediction of distortion. Once trained, the model can efficiently predict distortions under unseen manufacturing parameters. By combining validated simulations with operator-based learning, the framework supports both reliable distortion modelling and real-time prediction.

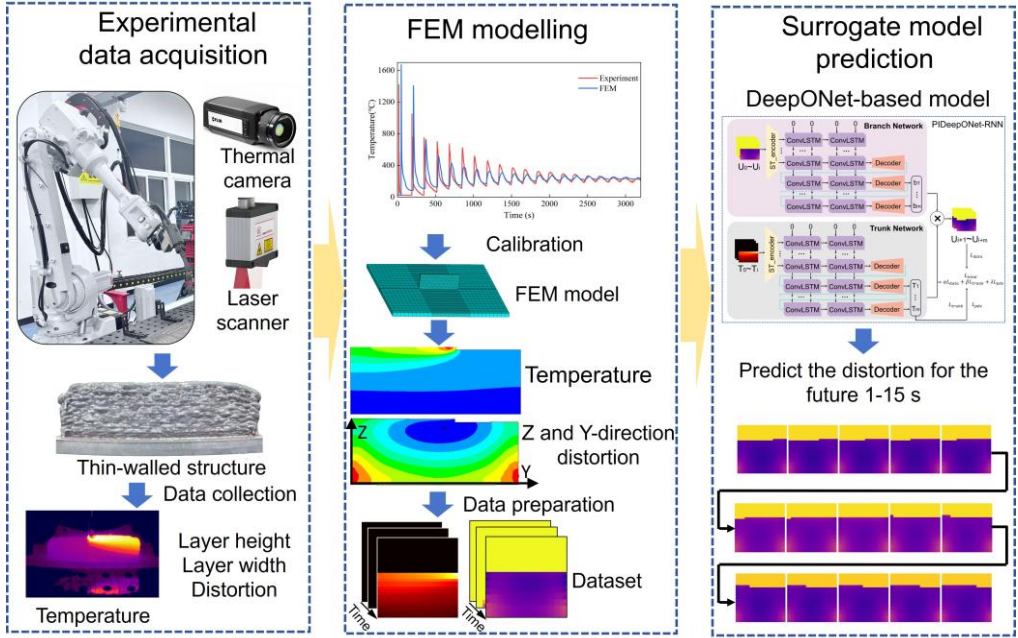

**Fig. 1. The overall framework of the proposed real-time distortion prediction system in metallic AM.**

2.2. FEM modelling and data preparation

To build the FEM datasets for validating the distortion prediction model, this study employs ANSYS APDL to perform thermo-mechanical coupled FEM simulations of the WAAM process, followed by extraction and preprocessing of relevant data to generate the datasets. The heat input follows the Goldak double-ellipsoidal heat-source model[41], whose shape parameters are calibrated based on existing studies[42] and experiment. Thin-wall material is ER70S-6 low-carbon steel wire, and substrate is Q235b low-carbon steel. Thin-walled structures are selected as the research subjects due to their extensive demand in aerospace lightweight applications, while also providing an important theoretical modeling and validation basis for AM process optimization and quality control of complex thin-walled structures[43].

Figure 2(a) shows the established FEM model of the thin-walled structure, including the meshing scheme and size settings. The model consists of a substrate measuring $300 \times 300 \times 10$ mm$^3$ and a thin-walled structure with height of 100 mm. The deposition layer width and height are determined based on experimental measurements and previous research[42]. Both the substrate and thin-walled structure are meshed with hexahedral elements. The mesh size is validated through a mesh independence study, ensuring computational accuracy while effectively controlling computational cost. The "birth-death element" technique is employed to simulate the layer-by-layer material deposition process, where elements are sequentially activated with corresponding thermal loads to achieve dynamic addition of deposited material. FEM employs a sequential thermo-mechanical coupling method, in which the temperature field obtained from solving the transient heat conduction equation is applied as a thermal load in the subsequent mechanical analysis to determine the resulting physics fields. Thermal input raises the temperature, leading to expansion that is resisted by external constraints, thereby generating stress and distortion. Fixed constraints are applied along the z-direction at the four corner points of the substrate. To avoid potential overheating in the arc initiation area, uneven heat distribution, and forming dimensional deviations caused by unidirectional deposition, this study employs a zigzag scanning path. A 60 s interlayer cooling time is set after each deposited layer to control interlayer heat accumulation. The simulations were executed on a personal laptop and required approximately 4 hours.

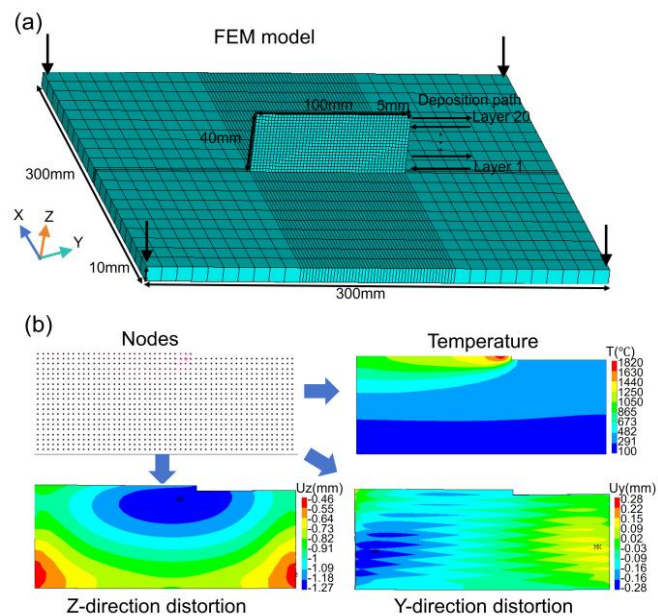

**Fig. 2. FEM modelling and visualization of multi-physics field results. (a)** FEM model with constraint and deposition path, **(b)** data collection and preprocessing.

FEM simulation obtained the results of the temperature field, z and y directions distortion, as shown in Figure 2(b). Due to the symmetrical features of the thin-walled structure on both sides of the y-z plane, data are extracted from nodes on one side surface with a spatial resolution of 2 mm. Following the FEM model design, the temperature in non-deposited areas is set to ambient temperature, and the distortion is set to zero. Data structure is a 2D matrix containing temperature, z and y-direction distortion. The temperature and distortion histories are recorded at 1 s sampling interval, ensuring both data sufficiency and temporal resolution adequate for accurate learning. Datasets are normalized to the range [0, 1].

2.3. DeepONet-based surrogate models for distortion prediction

To achieve efficient and generalizable modeling of the complex thermo-mechanical coupled distortion in WAAM, DeepONet-based surrogate model is proposed, whose architecture is illustrated in Figure 3. The model aims to accurately and rapidly predict the spatiotemporal dynamics of distortion field U, driven by the evolution of temperature field T, by learning from historical process data. It predicts the distortion $\hat{U}$ of the thin-walled structure during the deposition stage along the z and y-directions. The surrogate model captures the influence of thermal history on mechanical distortion, enabling rapid and scalable predictions across varying process parameters and geometric evolution, thereby replacing traditional time-consuming numerical simulations. The task addressed by the surrogate model can be defined as:

$$[\hat{U}_{i+1}, \hat{U}_{i+2}, \ldots, \hat{U}_{i+m}] = \mathcal{F}_\theta \left( \left\{ \begin{bmatrix} T_0 \\ U_0 \end{bmatrix}, \begin{bmatrix} T_1 \\ U_1 \end{bmatrix}, \ldots, \begin{bmatrix} T_i \\ U_i \end{bmatrix} \right\} \right) \quad (1)$$

It represents predicting distortion for the future $m$ time steps using historical temperature field and distortion data from the deposited time steps.

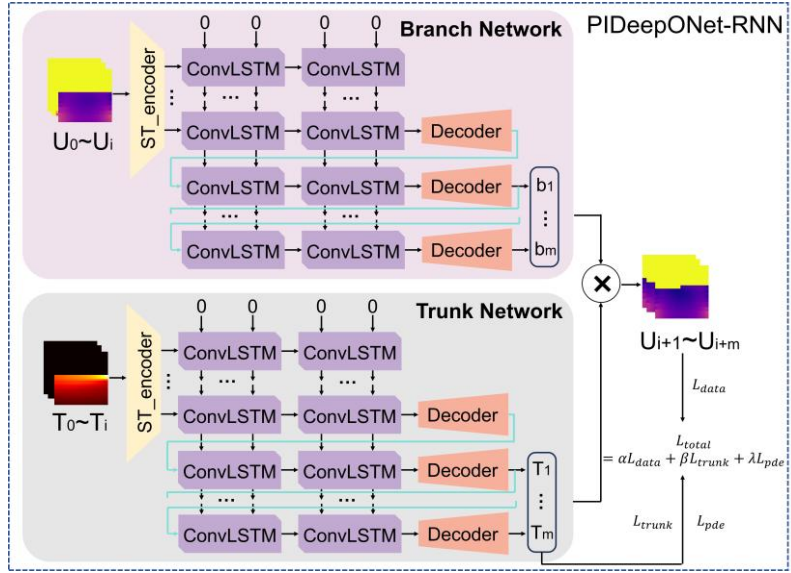

**Fig. 3. Architecture of the PIDeepONet-RNN surrogate model.**

Distortion of the WAAM part is not determined by its state at a single time instant, but rather results from the accumulated history of thermo-mechanical coupling[44]. The model is improved DeepONet-based to address the spatiotemporal features of temperature and distortion fields in WAAM, as well as the requirement for long-horizon prediction. It consists of a trunk network and a branch network, effectively decoupling the complex thermo-mechanical coupling problem. The trunk network takes the temperature history $T_{history} = [T_t]_{t=0}^i$ as input, where $i$ denotes the total historical time steps, $T_t \in \mathbb{R}^{H \times W}$, and $t \in \mathbb{R}^i$, to extract the evolutionary features of the driving load and thermal input. It outputs the predicted temperature field $t_k$ for the future $m$ s, where $k \in \mathbb{R}^m$. The branch network takes the distortion history $U_{history} = [U_t]_{t=0}^i$ as input, where $U_t \in \mathbb{R}^{H \times W}$. It encodes and extracts the encoded features that represent the physical properties and dynamic response patterns of the system itself, and outputs a feature map corresponding to future times. Finally, the outputs of the branch and trunk networks are coupled globally using a field-level Hadamard product to generate the final prediction of the future distortion field.

$$\hat{U} = \sigma(b_k \odot t_k) \quad (2)$$

Where $\odot$ is the Hadamard product, and $\sigma$ is the Sigmoid activate function. This operation

physically simulates the coupling process where future thermal loads act upon the mechanical state to produce the final response.

DeepONet itself does not rely on a recurrent structure. It is originally designed to handle mappings between function spaces, rather than sequential or time-dependent data. Its training and inference processes are feedforward, with no explicit recurrent connections. To handle time series or dynamic systems, DeepONet needs to be combined with an RNN. To achieve RNN-based functionality, both the trunk and branch networks adopt a "Spatiotemporal Encoder + Convolutional Long Short-term Memory (ConvLSTM)" architecture. Before entering the RNN, an encoder composed of 3D convolutional layers first processes the input long-horizon historical data. This encoder primarily performs convolution and down sampling along the temporal dimension, aiming to compress lengthy raw data into a latent feature sequence with higher information density and shorter length.

The RNN component of the model consists of multiple ConvLSTM layers. Unlike standard LSTM that processes 1D vectors, ConvLSTM employs convolutional operations within its gating units, enabling it to process time series while completely preserving the spatial topology of the data at each time step[45]. This module first reads the entire latent feature sequence during the encoding stage, compressing historical information into its internal state; then, during the decoding stage, it generates predictions for the future time steps in an autoregressive manner.

To prevent the model from overfitting purely to data and to ensure the physical plausibility of its predictions, the transient heat conduction equation is introduced as a soft constraint on the trunk network's output to guide the training process. A finite difference approximation is employed to discretize the derivative terms in the governing heat conduction equation. The temperature field predicted by the trunk network is required to satisfy this equation:

$$R_t(x_i, y_j) = \rho C_p \frac{\partial T}{\partial t} - k\nabla^2 T + \dot{Q} \tag{3}$$

Where $\rho$ is the density of the part material, $C_p$ is the specific heat capacity, and $\dot{Q}$ is the arc heat source term. The physical loss term $L_{pde}$ is defined as:

$$L_{pde} = \frac{1}{nmT}\sum_{i=1}^{n}\sum_{j=1}^{m}\sum_{t=1}^{T} R_t(x_i, y_j)^2 \tag{4}$$

This physical residual is integrated into the model's total loss function, forming a combined data and physics-driven total loss $L_{total}$ defined as:

$$L_{total} = \alpha L_{data} + \beta L_{trunk} + \lambda L_{pde} \tag{5}$$

Where $L_{data}$ is the Mean Squared Error (MSE) between the predicted distortion and the ground truth. $L_{trunk}$ is the MSE between the temperature predicted by the trunk network and the ground truth temperature. $L_{pde}$ is the MSE of the physical residual computed from the trunk network's output. $\alpha$, $\beta$, and $\lambda$ are weights used to balance the individual loss terms. By minimizing the total loss function $L_{total}$, the PIDeepONet-RNN is driven not only to fit the observed data but also to ensure that its internal simulation of the thermal process adheres to physical principles. This approach helps enhance the model's generalization capability and its predictive robustness in data-sparse or extrapolative areas. Additionally, the DeepONet-RNN (without the trunk output constraint) is used for comparative analysis and ablation studies in the results section.

The proposed surrogate model in this study uses the same spatiotemporal sequence encoding to regress the output future target and is trained in a supervised manner. For each sample of historical temperature and distortion fields, the model predicts future distortion and compares it with the ground truth using MSE loss. The Kullback-Leibler (KL) divergence is used to evaluate the difference between the predicted probability distribution and the ground truth probability distribution[42]. Both the ground truth temperature and predicted temperature are normalized into probability distributions, yielding the predicted probability distribution $Q_t$ and the ground truth probability distribution $P_t$.

$$Q_t(x_i, y_j) = \frac{\widehat{U}_t(x_i, y_j)}{\sum_{i'=1}^{n} \sum_{j'=1}^{m} \widehat{U}_t(x_{i'}, y_{j'})} \tag{6}$$

$$P_t(x_i, y_j) = \frac{U_t(x_i, y_j)}{\sum_{i'=1}^{n} \sum_{j'=1}^{m} U_t(x_{i'}, y_{j'})} \tag{7}$$

Therefore, by calculating the KL divergence between $Q_t$ and $P_t$, the distributional difference between the predicted distortion field and the ground truth distortion field is quantified. In this study, a smaller KL divergence value indicates a closer match between the distributions of the ground truth and the predicted distortion field.

$$D_{KL}(P_t \parallel Q_t) = \sum_{t=1}^{T} P_t \log \frac{P_t}{Q_t} \tag{8}$$

Additionally, the Structural Similarity Index (SSIM) is adopted as an evaluation metric for the prediction results. Unlike MSE or KL divergence, which only measure pixel-level errors, SSIM comprehensively assesses the structural pattern consistency between predicted and ground truth fields from three dimensions: luminance, contrast, and structure. It effectively evaluates the accuracy of structural features such as overall distortion trends, spatial distribution, and gradient variations[46]. Its evaluation criteria better align with human visual perception and engineering applications' focus on the "morphology" of distortion fields rather than mere numerical values, thereby avoiding structural prediction errors that might be obscured by traditional metrics.

The surrogate model is trained using the Adam optimizer with a fixed learning rate of 0.001 and a batch size of 16. Training is stopped after 5000 epochs. All models are trained on a single NVIDIA GeForce RTX 4050 GPU (6 GB memory), and the training process took approximately 2.5 hours. The surrogate model is implemented in Python using the PyTorch deep learning framework.

## 3. Practical experiments

This section provides a detailed description of the experimental design of this study, ensuring the reliability of the simulation data. Furthermore, the models required for ablation experiment have been designed.

### 3.1. Simulation

The experiments aimed to calibrate and validate the thermo-mechanical FEM simulation developed in this study. After calibrated, the FEM results are treated as the ground truth labels for training the surrogate models. The experimental system is set up based on previous studies[42]. Laser scanner is used to measure the final distortion, layer height, and layer width. This study selects 20 combinations of process parameters with different wire feed speeds and travel speeds for FEM modeling of thin-walled structures, the specific parameters are listed in Table 1. The layer height and width in the FEM model are calibrated using data from a laser scanner. The temperature and distortion data collected from experiments are used to calibrate the thermal input of the model, ensuring that the simulation values strike a balance between the errors in actual temperature and distortion. In total, Data from all 20 simulation cases are collected, resulting in 6,880 individual datasets.

Table 1 Process parameters for the thermo-mechanical FEM of WAAM

|  | WFS(m/min) | TS(m/min) | Height(mm) | Width(mm) |
| --- | --- | --- | --- | --- |
| Case 1 | 5 | 0.48 | 2 | 5 |
| Case 2 | 6 | 0.48 | 2 | 6 |
| Case 3 | 6 | 0.4 | 3 | 8 |
| Case 4 | 5 | 0.4 | 2.5 | 8 |
| Case 5 | 5 | 0.45 | 2.5 | 6 |
| Case 6 | 7 | 0.3 | 4.5 | 8 |
| Case 7 | 7 | 0.2 | 4.5 | 8 |
| Case 8 | 4 | 0.4 | 2 | 5 |
| Case 9 | 8 | 0.6 | 2 | 8 |
| Case 10 | 6 | 0.2 | 4 | 8 |
| Case 11 | 5 | 0.2 | 4 | 10 |
| Case 12 | 6 | 0.6 | 3 | 8 |
| Case 13 | 6 | 0.3 | 4 | 8.5 |
| Case 14 | 7 | 0.4 | 4 | 8 |
| Case 15 | 4 | 0.6 | 3.5 | 7 |
| Case 16 | 8 | 0.3 | 3 | 9 |
| Case 17 | 5 | 0.3 | 3 | 8.5 |
| Case 18 | 8 | 0.4 | 3 | 8 |
| Case 19 | 5.5 | 0.3 | 3 | 6 |
| Case 20 | 5 | 0.6 | 4 | 7 |

The complete training datasets comprises 16 thin-walled structures with varying process parameters. This ensures that the surrogate model does not overfit to a narrow range of geometric evolution scenarios, but instead demonstrates robust generalization under WAAM conditions with diverse process parameters and evolving geometric morphologies. These samples capture diverse historical data and physical field distribution configurations while maintaining the same spatiotemporal grid. Due to variations in process parameters, the printing time and geometric morphology of each layer differ. To evaluate generalization to unseen process parameters and geometric evolution, the model is evaluated on four independent datasets. The test datasets contain a total of 1,300 samples.

3.2. Ablation experiment

To explore the coupling relationship between thermal history and mechanical response in WAAM, two conventional data-driven surrogate models are proposed: CNN and ST-ConvLSTM model, as shown in Figure 4.

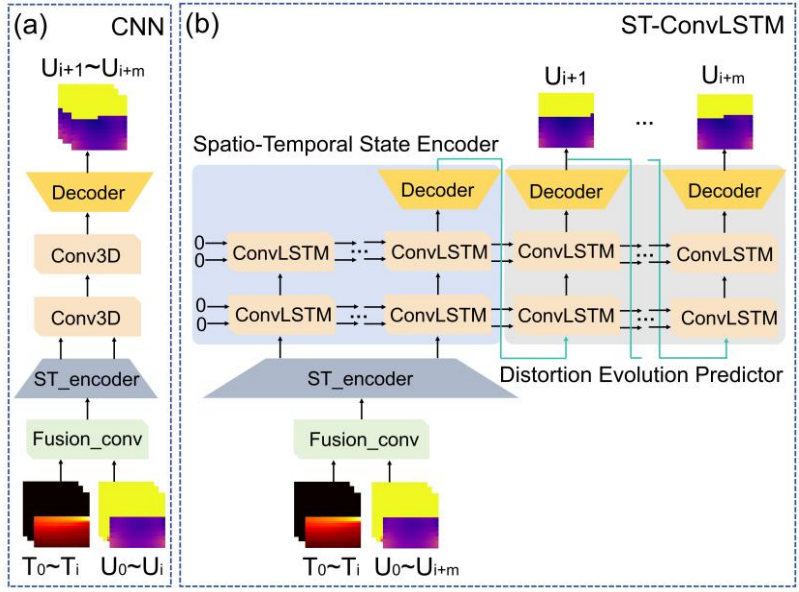

**Fig. 4. Conventional data-driven surrogated models. (a)** CNN. **(b)** ST-ConvLSTM.

CNN model is a direct mapping model. In this architecture, the input tensor first undergoes channel concatenation and preliminary feature fusion before entering the spatiotemporal encoder. This encoder consists of a series of 3D convolutional layers that perform stride convolutions along the temporal dimension, compressing lengthy temporal information and extracting dynamic features to generate a compact latent spatiotemporal representation. This representation is then fed into a deeper CNN module for hierarchical feature extraction, and finally, the output convolutional layer generates predictions for the entire future distortion sequence at once in a feedforward manner. The entire architecture possesses inherent parallel computing capabilities and the ability to capture local spatiotemporal features.

Spatiotemporal-ConvLSTM (ST-ConvLSTM) model introduces a recurrent mechanism, employing ConvLSTM units as its core. This model shares the same early coupling and spatiotemporal encoding front-end as the CNN model, but differs fundamentally in its processing of latent features. It is implemented based on an encoder-decoder architecture: the ConvLSTM network in the encoder sequentially processes the latent feature sequence, encoding all historical information into the final hidden state; the decoder then uses this hidden state as the initial condition to generate the future distortion time steps in an autoregressive manner, meaning each prediction is based on the output state from the previous times. This architecture can explicitly model long-horizon dependencies and causal relationships within the time series.

The fundamental distinction between these two types of coupled models lies in their core mechanisms for processing temporal information. CNN model performs direct spatiotemporal mapping, whereas ConvLSTM model conducts recurrent, autoregressive temporal generation. They will serve as critical benchmarks for evaluating the effectiveness of different feature fusion and temporal modeling strategies, and will be compared with the DeepONet-based decoupled model in ablation experiment.

## 4. Result and Discussion

In this section, the z and y-direction distortions predicted by the four surrogate models are compared with the FEM results in terms of model accuracy, error distribution, and time efficiency in long-horizon prediction. Furthermore, ablation studies are conducted to analyze the training loss curves, as well as the MSE, KL divergence, and SSIM curves for long-horizon prediction, in order to evaluate the model performance and the contribution of its components.

4.1. Z-direction distortion prediction

This section evaluates the z-direction distortion prediction performance for the future 1-15 s of the four surrogate models for thin-walled structure with unseen process parameters over cross-layer long-horizon, using case 20 at 1192 s of deposition as an example.

4.1.1. Comparison of temporal error distribution in long-horizon prediction

To evaluate the z-direction distortion prediction accuracy of the proposed surrogate models, the predictions from four surrogated models are compared with the FEM results. Figures 5-7 show the z-direction distortion predictions for the future 1-5 s, 6-10 s, and 11-15 s respectively, comparing the predicted results of the CNN, ST-ConvLSTM, DeepONet-RNN, and PIDeepONet-RNN models with the FEM results. The visualizations include the true distortion field from the FEM result, the corresponding model predictions, and the overall and the Region of Interest (ROI) area 3D absolute prediction error. The ROI is located at the corner of the deposition path within the thin-walled structure. The abrupt change in geometry induces elevated levels of strain and stress in this region, making it typically prone to cracking. Therefore, this study discusses the prediction results for this specific area. The study primarily focuses on how accuracy varies with the prediction horizon. For quantitative performance evaluation, Table 2 summarizes the average Mean Absolute Error (MAE) and maximum absolute error for the predicted results of each time segment.

**Table 2** Prediction performance comparison of four surrogate models for z-direction distortion in the future 1-5 s, 6-10 s, and 11-15 s.

|  | Times | Average MAE (mm) | Max absolute error (mm) |
|---|---|---|---|
| CNN | 1-5 s | 0.5609 | 1.2239 |
|  | 6-10 s | 0.3575 | 1.2202 |
|  | 10-15 s | 0.0715 | 1.1903 |
| ST-ConvLSTM | 1-5 s | 0.0573 | 1.2875 |
|  | 6-10 s | 0.0541 | 1.1963 |
|  | 10-15 s | 0.0581 | 1.1781 |
| DeepONet-RNN | 1-5 s | 0.0323 | 1.1267 |
|  | 6-10 s | 0.0334 | 1.1884 |
|  | 10-15 s | 0.0445 | 1.1765 |
| PIDeepONet-RNN | 1-5 s | 0.0261 | 0.9733 |
|  | 6-10 s | 0.0287 | 0.9393 |
|  | 10-15 s | 0.0346 | 1.1662 |

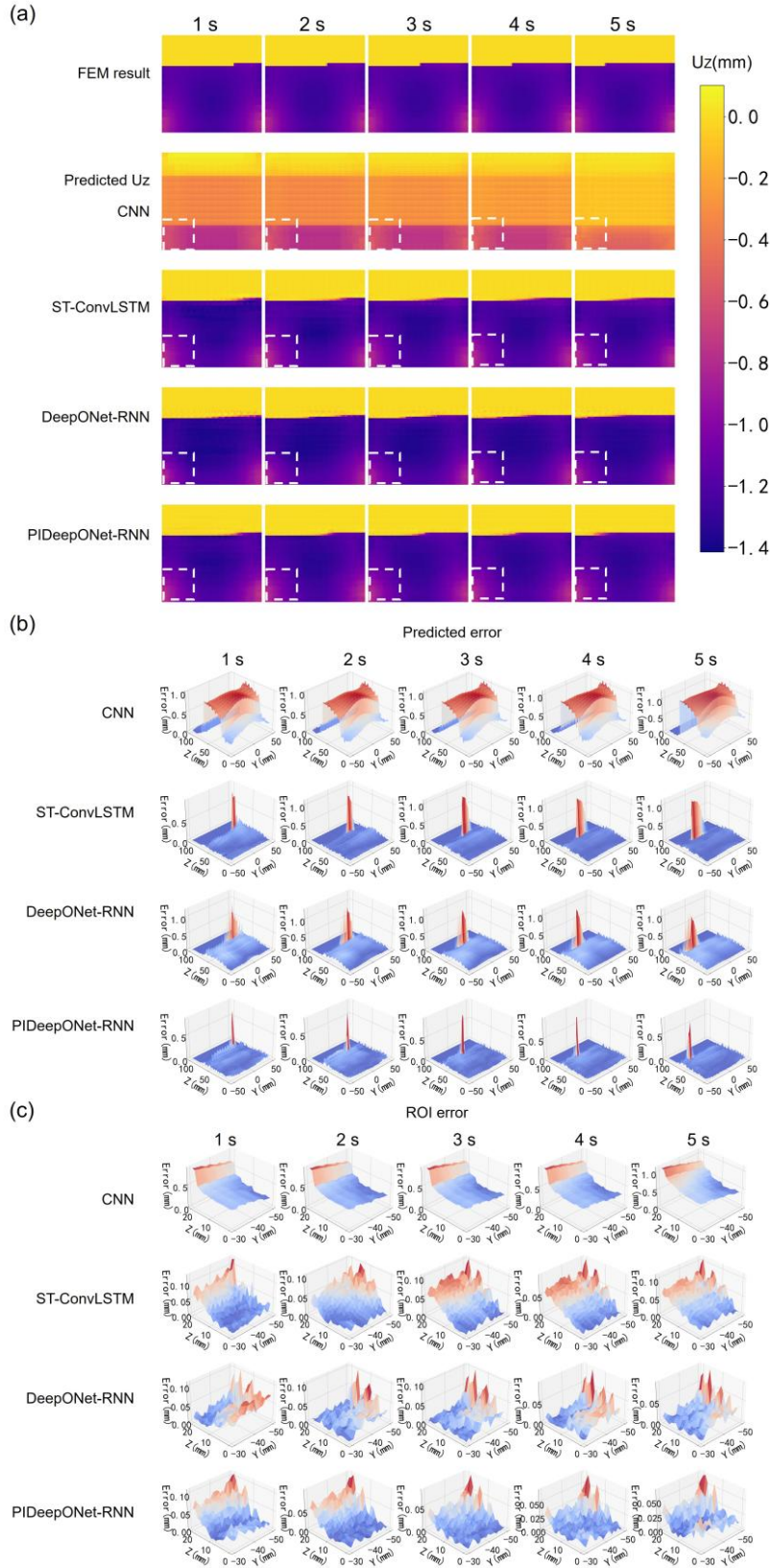

**Fig. 5. Performance of surrogate models for z-direction distortion prediction at 1192 s of deposition for the future 1-5 s. (a)** Comparison of the prediction results with FEM results. **(b)** Comparison of predicted distortion fields absolute errors. **(c)** Comparison of predicted distortion fields ROI absolute errors.

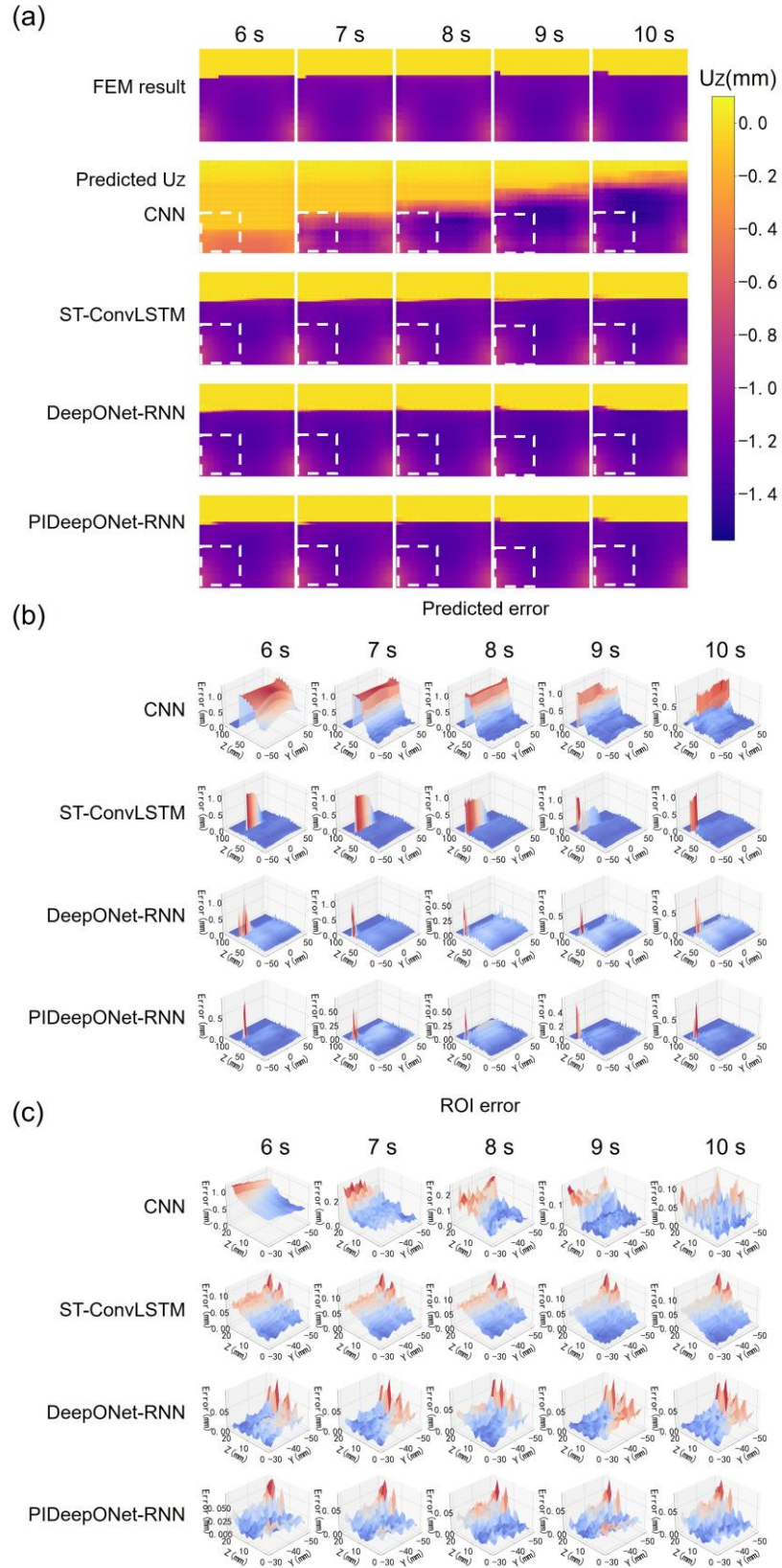

**Fig. 6. Performance of surrogate models for z-direction distortion prediction at 1192 s of deposition for the future 6-10 s. (a)** Comparison of the prediction results with FEM results. **(b)** Comparison of predicted distortion field. **(c)** Comparison of predicted distortion fields ROI absolute errors.

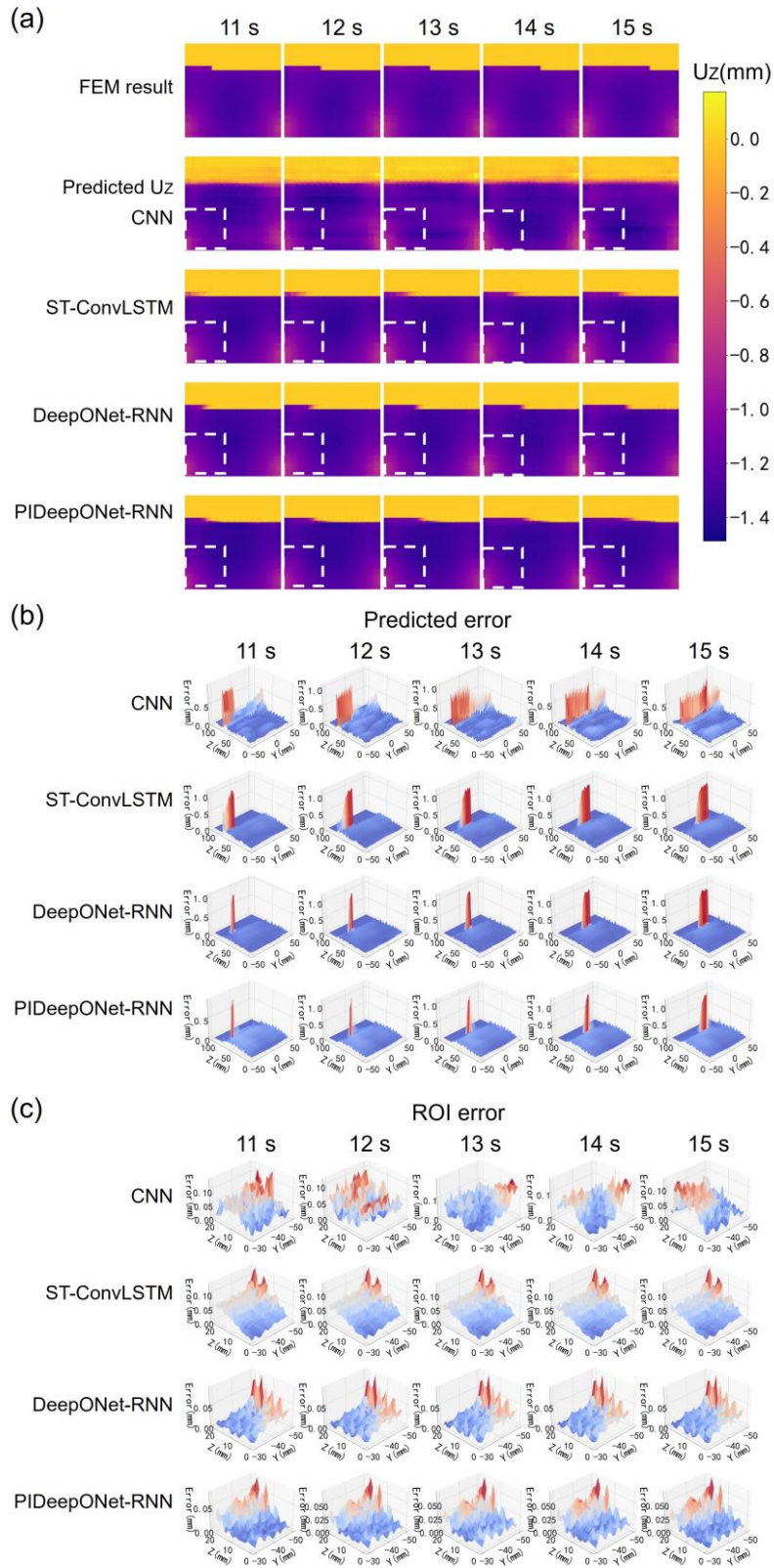

**Fig. 7. Performance of surrogate models for z-direction distortion prediction at 1192 s of deposition for the future 11-15 s. (a)** Comparison of the prediction results with FEM results. **(b)** Comparison of predicted distortion fields absolute error. **(c)** Comparison of predicted distortion fields ROI absolute errors.

Figures 5-7(a) illustrate the visual difference between the model predictions and FEM results. Among all models, the CNN shows the weakest performance, failing to capture the overall distortion distribution. In contrast, the other surrogate models provide more accurate representations, with the main discrepancies appearing in the molten pool region, as shown in Figures 5-7(b). Figures 5-7(c) further compare the ROI prediction errors. CNN model exhibits errors nearly an order of magnitude larger than the other models during the early stage (1-7 s). Although these errors decrease in the later stage (8-15 s), their spatial distribution remains irregular, with random errors mainly occurring at peripheral areas. By comparison, the other models show errors concentrated along the interior area, where the PIDeepONet-RNN achieves the lowest prediction error. Accurate prediction in the ROI is critical for mitigating distortion-related defects. CNN model exhibited the largest prediction deviation, with average MAE of 0.5609 mm, 0.3575 mm, and 0.0715 mm for the three intervals of 1-5 s, 6-10 s, and 11-15 s, respectively, and a max absolute error of 1.2239 mm. Although its later predictions show gradual improvement, it's also inability to model the distortion evolution during layer accumulation arises from its reliance on local convolution operations, which limits its capacity to capture long-range spatiotemporal dependencies between temperature and distortion fields.

In contrast, the ST-ConvLSTM model achieves relatively accurate predictions of distortion distribution in the deposited layers. This model effectively captures spatiotemporal correlations in the historical data and enhances long-horizon feature extraction. Its average MAE for the three time periods is 0.0573 mm, 0.0541 mm, and 0.0581 mm, respectively, representing a reduction of about 30% compared to the CNN model, with relatively stable errors over time. However, its max absolute error remains as high as 1.2875 mm, indicating that while the model can learn certain spatiotemporal laws, its generalization and robustness are limited when exposed to unseen dynamic conditions. The corresponding errors show that primarily concentrated around the molten pool area, while errors in already deposited areas approach zero.

DeepONet-RNN model, based on a decoupled operator learning framework, further enhances prediction accuracy. It achieves average MAEs of 0.0323 mm, 0.0334 mm, and 0.0445 mm, with both the magnitude and spatial extent of errors reduced. This improvement stems from its decoupling strategy, which separates the learning of thermal dynamics and mechanical responses, allowing for more efficient internal representations. Nevertheless, as a purely data-driven model, it still exhibits minor unphysical local biases, particularly around geometric features and welding boundaries. By introducing physics-based constraints derived from the heat conduction equation into the loss function, the PIDeepONet-RNN achieves the best overall performance. Its maximum absolute error decreases to 0.9733 mm, and the average MAE for the 1–5 s interval is reduced to 0.0261 mm. The inclusion of physical constraints serves as an effective regularization mechanism, restricting the solution space to physically consistent behaviors. As a result, spurious predictions in data-sparse or highly nonlinear regions, such as near the heat source, are largely suppressed. Errors in the molten pool area become negligible, which is acceptable given the limited measurement accuracy and low practical relevance of this region.

4.1.2. Comparison of spatial error distribution in long-horizon prediction

To evaluate the z-direction distortion error distribution of the proposed surrogate models in long-horizon prediction, the predictions from four surrogated models are compared with the FEM results. The study primarily focuses on how spatial error distribution in long-horizon prediction.

For quantitative performance evaluation, Table 3 summarizes the gradient magnitude for the overall, molten pool area, deposited area, and ROI of the predicted results error for each time segment.

Table 3 Gradient norm magnitude comparison of four surrogate models for z-direction distortion in the future 1-5 s, 6-10 s, and 11-15 s.

|  | Times | Overall | Molten pool area | Deposited area | ROI |
|---|---|---|---|---|---|
| CNN | 1-5 s | 0.0352 | 0.0769 | 0.028 | 0.036 |
|  | 6-10 s | 0.0414 | 0.147 | 0.0322 | 0.0245 |
|  | 10-15 s | 0.0338 | 0.1498 | 0.0184 | 0.0204 |
| ST-ConvLSTM | 1-5 s | 0.0151 | 0.0722 | 0.0094 | 0.0097 |
|  | 6-10 s | 0.0165 | 0.1631 | 0.0077 | 0.0092 |
|  | 10-15 s | 0.0169 | 0.1045 | 0.0077 | 0.0097 |
| DeepONet-RNN | 1-5 s | 0.013 | 0.071 | 0.0089 | 0.0087 |
|  | 6-10 s | 0.0116 | 0.0708 | 0.0082 | 0.0088 |
|  | 10-15 s | 0.014 | 0.0737 | 0.0082 | 0.0091 |
| PIDeepONet-RNN | 1-5 s | 0.01 | 0.0407 | 0.0081 | 0.0083 |
|  | 6-10 s | 0.0095 | 0.0441 | 0.0082 | 0.0083 |
|  | 10-15 s | 0.0132 | 0.0733 | 0.0077 | 0.008 |

As shown in Table 3, the gradient norms of all four surrogate models are higher in the molten pool area and lower in the deposited area, indicating that errors are primarily concentrated in the molten pool area. The ROI is slightly higher than those in the deposited area. During the early prediction stage (1-5 s), the CNN model produces widely distributed errors across the computational domain. The gradient norm in the molten pool area exceeds that in the ROI and deposited regions, reflecting its limited ability to capture long-range spatiotemporal dependencies. This limitation arises from its direct spatiotemporal mapping and restricted receptive field. As the prediction horizon increases (6-15 s), the errors gradually converge toward the active molten pool and welding boundaries, with the gradient norm reaching 0.1498, higher than that of the overall field. This shift occurs because the distortion patterns in solidified regions become more uniform over time, requiring long-term temporal information to model accurately. However, the CNN lacks a temporal memory mechanism, leading to cumulative errors when handling dynamic molten pool evolution.

ST-ConvLSTM model can effectively captures temporal dependencies in the deposited areas during the early stage. Its initial errors are concentrated around the molten pool, with a gradient norm of 0.0722, about 7 times larger than that of the deposited area. However, as the prediction horizon extends, the model exhibits error accumulation, particularly during dynamic interlayer transitions. At 11-15 s, the overall and molten pool gradient norms reach 0.0169 and 0.1045, respectively, indicating a broader error spread. This is primarily due to the autoregressive nature of prediction, where small deviations are propagated and amplified over time, exposing the model's limited adaptability to geometric variations.

DeepONet-RNN model achieves better long-horizon stability. Its initial errors are similarly concentrated in the molten pool area but with smaller magnitudes. The gradient norm is 0.071, which is lower than that of the ST-ConvLSTM model. Nevertheless, as prediction proceeds, it still exhibits gradual error accumulation trend of purely data-driven methods.

PIDeepONet-RNN model achieves the best overall performance, showing minimal initial errors concentrated near the molten pool. Its gradient norm in this area reaches the minimum value of 0.0407 among all models, and the error growth remains smallest throughout the 15 s prediction horizon. Even during the transition from layer 19 to layer 20, where geometric and boundary conditions change, the model maintains stable error distributions. The embedded heat conduction equation serves as a strong physical regularization, constraining the solution space to a physically consistent manifold. Consequently, the model not only learns statistical correlations from data but also captures the underlying physical principles, effectively suppressing cumulative errors and enhancing its generalization across diverse process conditions and geometric evolutions.

4.2. Y-direction distortion prediction

This section evaluates the y-direction distortion prediction performance for the future 1-15 s of the four surrogate models for thin-walled structure with unseen process parameters over cross-layer long-horizon, using case 20 at 1192 s of deposition as an example.

4.2.1. Comparison of temporal error distribution in long-horizon prediction

To further analyzes the y-direction distortion prediction accuracy of the proposed surrogate models, the predictions from four surrogated models are compared with the FEM results. Figures 8-10 shows the y-direction distortion predictions for the future 1-5 s, 6-10 s, and 11-15 s respectively, comparing the predicted results of the CNN, ST-ConvLSTM, DeepONet-RNN, and PIDeepONet-RNN models with the FEM results. The visualizations include the distortion field from the FEM result, the corresponding model predictions, and the 3D overall and ROI absolute prediction error. For quantitative performance evaluation, Table 4 summarizes the average MAE and max absolute error for the predicted results of each time segment.

**Table 4** Prediction performance comparison of four surrogate models for y-direction distortion in the future 1-5 s, 6-10 s, and 11-15 s.

|  | Times | Average MAE (mm) | Max absolute error (mm) |
| --- | --- | --- | --- |
| CNN | 1-5 s | 0.1317 | 0.4917 |
|  | 6-10 s | 0.0752 | 0.4914 |
|  | 10-15 s | 0.0318 | 0.4911 |
| ST-ConvLSTM | 1-5 s | 0.0172 | 0.2361 |
|  | 6-10 s | 0.0162 | 0.3334 |
|  | 10-15 s | 0.0196 | 0.3506 |
| DeepONet-RNN | 1-5 s | 0.0168 | 0.2223 |
|  | 6-10 s | 0.0152 | 0.2972 |
|  | 10-15 s | 0.0195 | 0.3478 |
| PIDeepONet-RNN | 1-5 s | 0.0165 | 0.2049 |
|  | 6-10 s | 0.0149 | 0.2125 |
|  | 10-15 s | 0.0157 | 0.3288 |

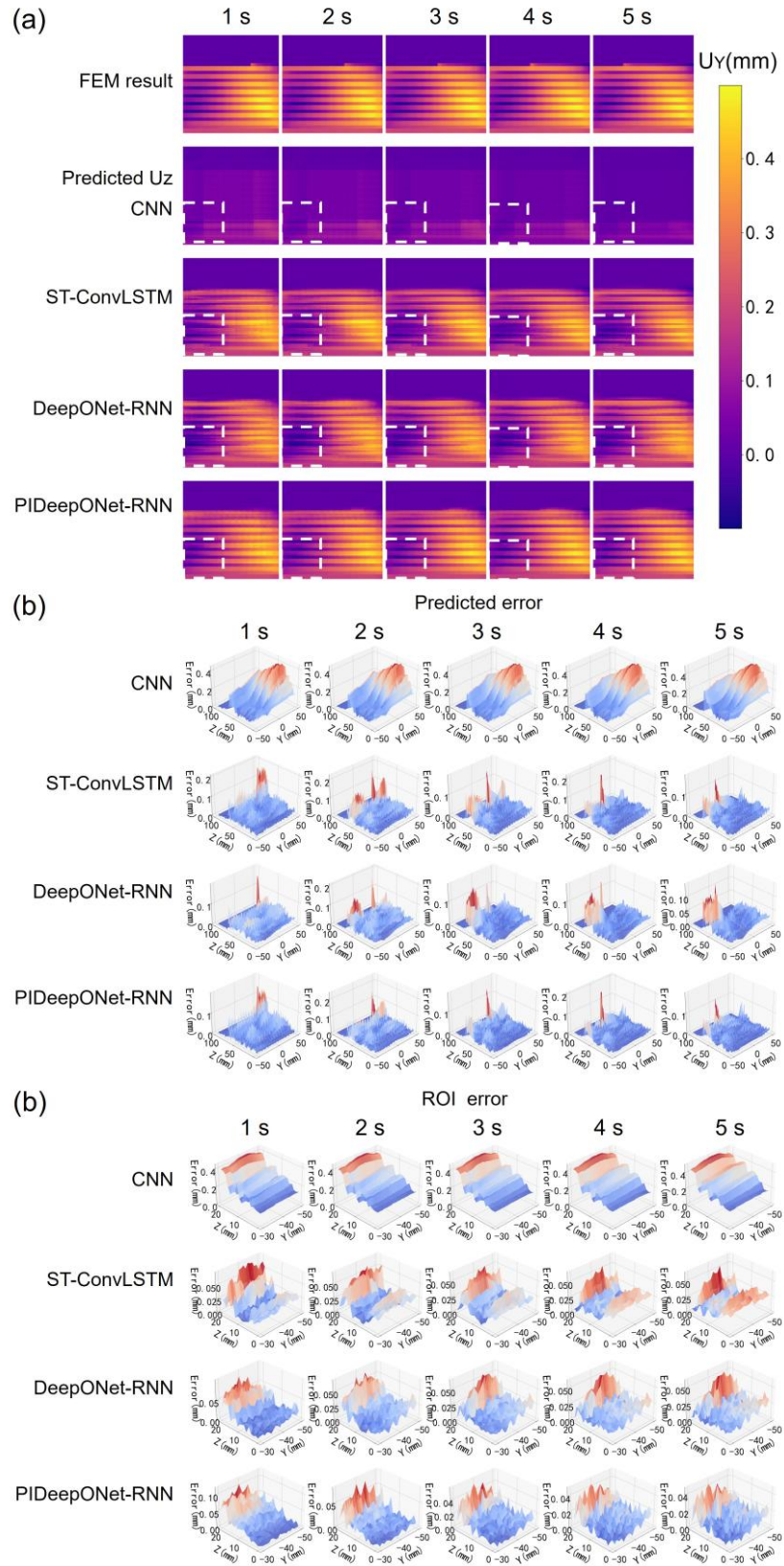

**Fig. 8. Performance of surrogate models for y-direction distortion prediction at 1192 s of deposition for the future 1-5 s. (a)** Comparison of the prediction results with FEM results. **(b)** Comparison of predicted distortion fields absolute errors. **(c)** Comparison of predicted distortion fields ROI absolute errors.

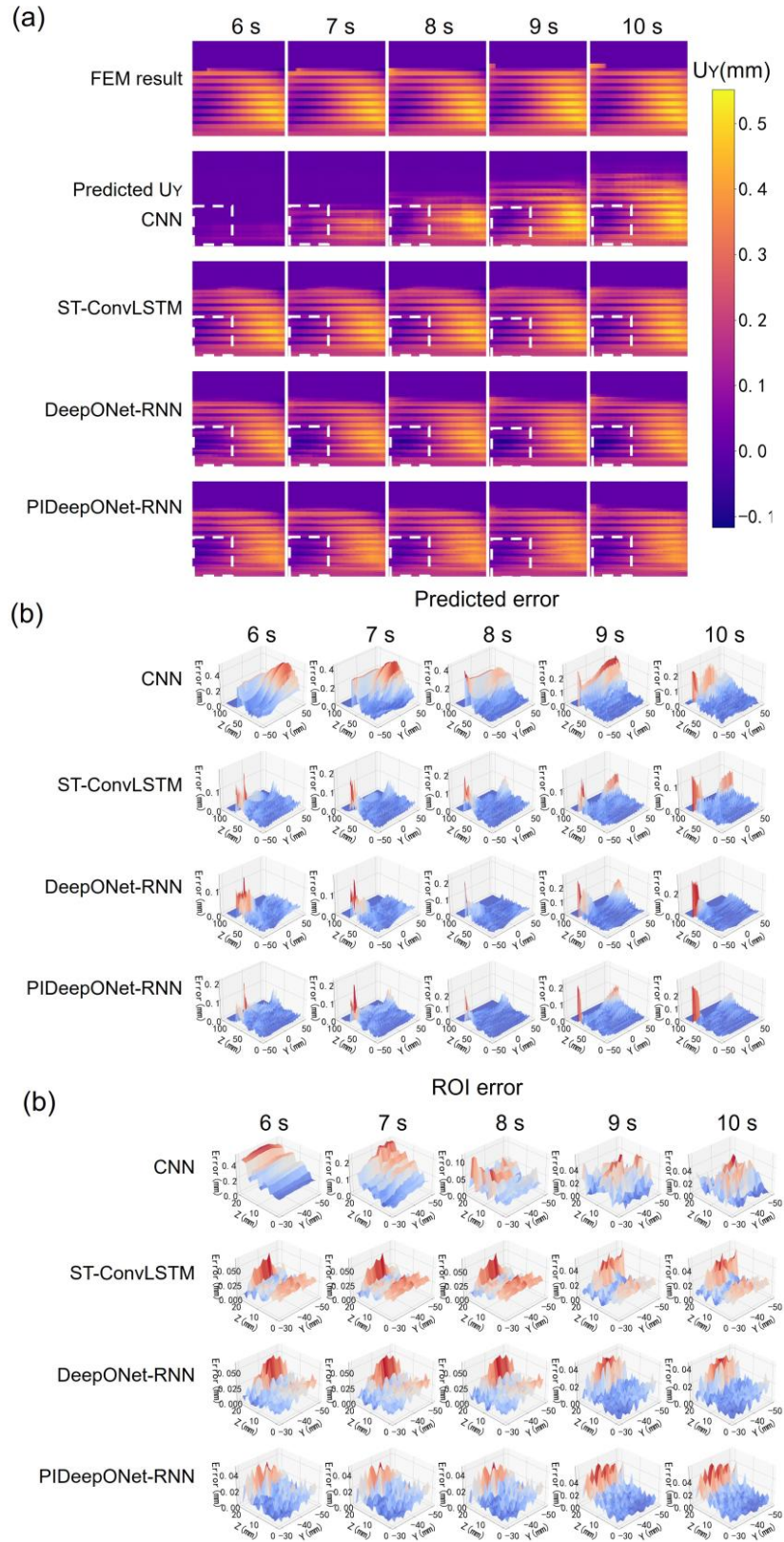

**Fig. 9. Performance of surrogate models for y-direction distortion prediction at 1192 s of deposition for the future 6-10 s. (a)** Comparison of the prediction results with FEM results. **(b)** Comparison of predicted distortion fields absolute errors. **(c)** Comparison of predicted distortion fields ROI absolute errors.

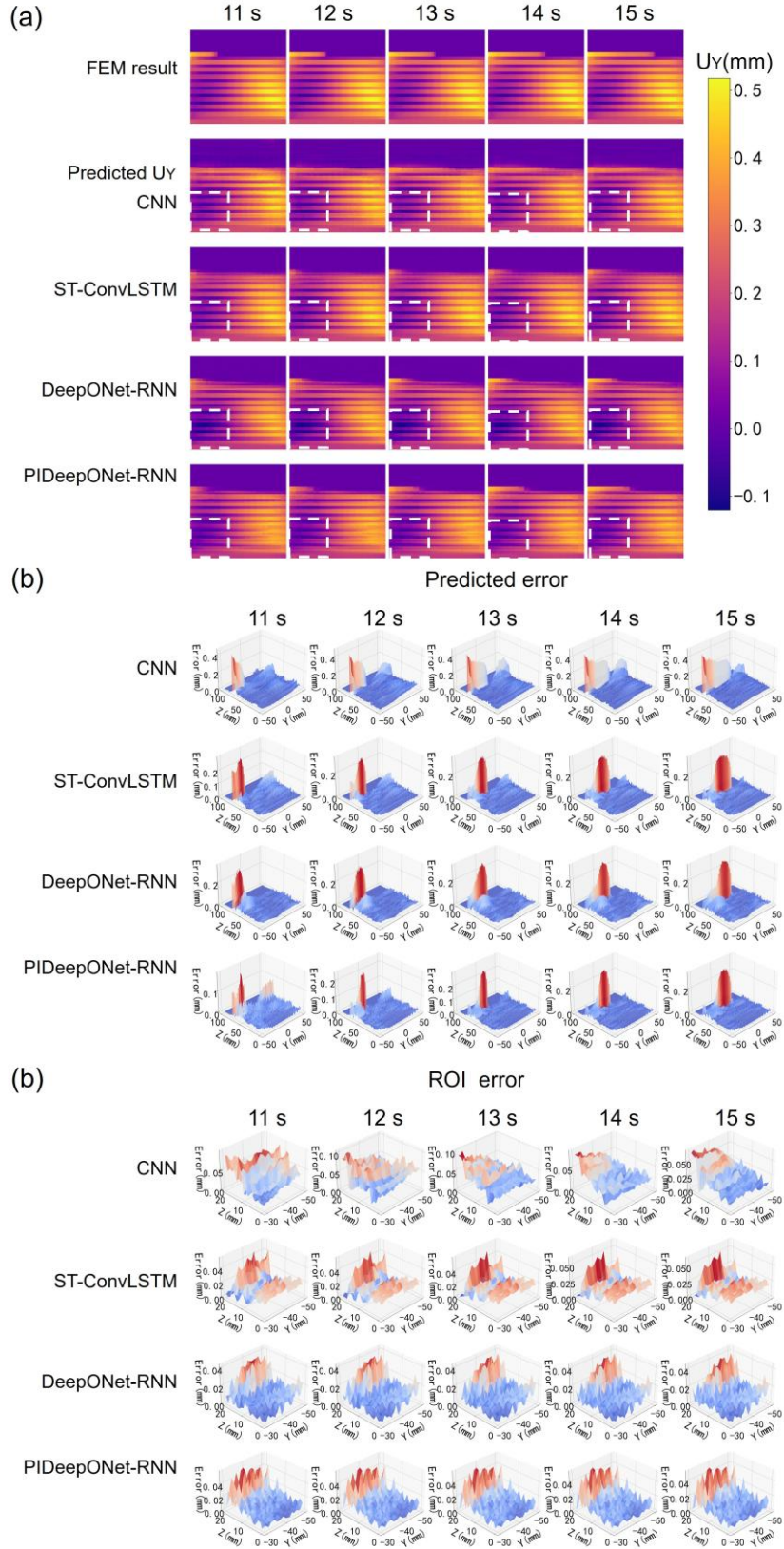

**Fig. 10 Performance of surrogate models for y-direction distortion prediction at 1192 s of deposition for the future 11-15 s. (a)** Comparison of the prediction results with FEM results. **(b)** Comparison of predicted distortion fields absolute errors. **(c)** Comparison of predicted distortion fields ROI absolute errors.

PIDeepONet-RNN model achieves the best performance, with average MAE of 0.0165 mm, 0.0149 mm, and 0.0157 mm for the three-time intervals (1-5 s, 6-10 s, and 11-15 s), respectively, and a max absolute error of only 0.3288 mm. In contrast, CNN model shows the largest prediction deviation, with average MAE of 0.1317 mm, 0.0752 mm, and 0.0318 mm, respectively. Its ROI absolute error during the early prediction stage is nearly an order of magnitude greater than that of the other models, though it becomes comparable in the later phase. Temporally, the CNN errors are concentrated in the early stage, whereas the other models show mild error accumulation over time. Both ST-ConvLSTM and DeepONet-RNN outperform the CNN by maintaining average MAE around 0.02 mm, demonstrating stronger temporal stability. However, they still produce localized high error area. The ST-ConvLSTM reaches a max absolute error of 0.3506 mm at 11-15 s. The max absolute error of DeepONet-RNN remains 0.3478 mm, revealing the limitations of purely data-driven frameworks in constraining unphysical distortions.

The average MAE of PIDeepONet-RNN reduced by about 10% compared to other models, with a more concentrated and physically reasonable spatial distribution. This indicates that the physical information effectively regularizes the network's solution space and suppresses structural responses that violate thermodynamic principles. The y-direction distortion prediction results further validate the previous conclusion. It should be noted that zigzag scanning path introduces asymmetric heat input and shear stresses along the y-direction. Thin-walled structure exhibits relatively low stiffness in the y-direction, making it incapable of resisting thermally induced distortion. This results in various distortion laws in the y-direction, including warping, distortion, and striation-like distortion. This phenomenon results in a striated error distribution with relatively low magnitude in the model's predictions for already deposited areas.

4.2.2. Comparison of spatial error distribution in long-horizon prediction

To further evaluate the y-direction distortion error distribution of the proposed surrogate models in long-horizon prediction, the predictions from four surrogated models are compared with the FEM results. For quantitative performance evaluation, Table 5 summarizes the gradient magnitude for the overall, molten pool area, deposited area, and ROI of the predicted results error for each time segment.

**Table 5** Gradient norm magnitude comparison of four surrogate models for y-direction distortion in the future 1-5 s, 6-10 s, and 11-15 s.

|  | Times | Overall | Molten pool area | Deposited area | ROI |
|---|---|---|---|---|---|
| CNN | 1-5 s | 0.0373 | 0.0812 | 0.0487 | 0.0444 |
|  | 6-10 s | 0.0249 | 0.0511 | 0.0293 | 0.022 |
|  | 10-15 s | 0.0149 | 0.0512 | 0.0107 | 0.0131 |
| ST-ConvLSTM | 1-5 s | 0.0088 | 0.0265 | 0.0093 | 0.0087 |
|  | 6-10 s | 0.0071 | 0.0251 | 0.0078 | 0.0074 |
|  | 10-15 s | 0.01 | 0.0369 | 0.0081 | 0.0069 |
| DeepONet-RNN | 1-5 s | 0.0085 | 0.0242 | 0.0092 | 0.0084 |
|  | 6-10 s | 0.0074 | 0.0216 | 0.0077 | 0.0067 |
|  | 10-15 s | 0.0093 | 0.033 | 0.0081 | 0.0064 |
| PIDeepONet-RNN | 1-5 s | 0.008 | 0.0148 | 0.0093 | 0.0077 |

| | | | | |
|---|---|---|---|---|
| 6-10 s | 0.0072 | 0.021 | 0.0075 | 0.0064 |
| 10-15 s | 0.0086 | 0.0283 | 0.0075 | 0.0063 |

For the CNN model, during the early prediction stage (1-5 s), errors are widely distributed across already deposited areas. Unlike the gradient norm of z-direction distortion error, the y-direction distortion error shows gradient norms of 0.0487 in deposited area and 0.373 in the overall, with deposited area being slightly larger than the overall. The gradient norm of ROI demonstrates a spatially more accurate prediction over time, decreasing from 0.0444 to 0.0131. In contrast to the gradient norm for the distortion predicted in the z-direction, the ROI in the y-direction exhibits a lower gradient norm relative to that of the deposited area. As the prediction horizon expands, although the gradient norm of prediction error in deposited area decreases to 0.107, this does not reflect genuine accuracy improvement but results from the model consistently generating smooth predictions lacking dynamic details. Errors gradually shift toward welding boundaries and the molten pool area.

Both the ST-ConvLSTM and DeepONet-RNN models also exhibit error accumulation. As prediction progresses, gradient norm continuously increases, where errors accumulated at welding boundaries become pronounced. This time-dependent error growth trend clearly reflects how purely data-driven models experience gradual degradation in generalization capability during long-horizon autoregressive inference due to accumulated minor deviations, causing predictions to progressively deviate from the true physical state.

In contrast, PIDeepONet-RNN model effectively mitigates the error accumulation problem. Throughout the entire 15 s prediction horizon, gradual degradation shows relatively small and no sustained increasing trend. This advantage is specifically manifested in the model ability to produce more accurate and physically plausible predictions at weld boundary areas even during later stages of long-horizon prediction, whereas purely data-driven models are more prone to generating unphysical artifacts in these areas.

4.3. Ablation study

This section conducts a ablation study on the four surrogate models for distortion prediction in both z and y-directions, aiming to evaluate the individual contributions of the network architecture, decoupling framework, and physical information constraints. The study employs four key metrics for quantitative analysis: training loss, MSE, KL divergence, and SSIM, as shown in Figure 11 and 12.

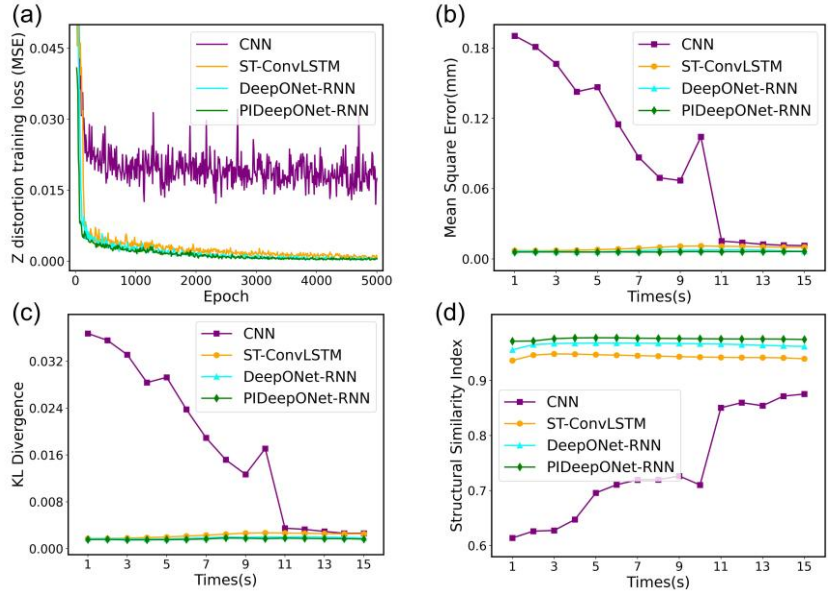

**Fig. 11. Evaluation of surrogate models for z-direction distortion prediction for the future 1-15 s. (a)** Comparison of the training loss. **(b)** Comparison of the MSE. **(c)** Comparison of the KL divergence. **(d)** Comparison of the SSIM.

Figure 11 presents the evaluation results of the four surrogate models for predicting z-direction distortion for the future 1-15 s. Figure 11(a) shows the training loss curves of each model. CNN model exhibits the highest loss values and fails to converge effectively, which indicates its limited ability to capture the complex spatiotemporal dynamics of the WAAM process. Both ST-ConvLSTM and DeepONet-RNN demonstrate steadily decreasing convergence trends, while PIDeepONet-RNN shows the smoothest convergence trajectory, suggesting that physical constraints optimize the training process. Figure 11(b) records the MSE variation across the prediction horizon. The CNN model shows the highest MSE values with fluctuations, confirming its poor temporal reliability. The MSE curves of ST-ConvLSTM and DeepONet-RNN are relatively close, but DeepONet-RNN maintains lower error levels. PIDeepONet-RNN consistently achieves the lowest MSE, with its advantage becoming more pronounced during the later prediction stage (11-15 s), demonstrating the effectiveness of physical constraints. Figure 11(c) quantifies the difference in probability distributions between predicted and ground truth fields using KL divergence. The CNN model shows the highest KL values, indicating the largest deviation between its predicted physical distributions and the actual situation. ST-ConvLSTM and DeepONet-RNN show reduced KL values with similar trends compared to CNN. PIDeepONet-RNN maintains the most stable and lowest KL values, proving that its predicted field distributions are closer to the real physical fields. Figure 11(d) evaluates prediction fidelity from the perspective of structural features using the SSIM index. PIDeepONet-RNN achieves SSIM values closest to 1, indicating that its predicted distortion fields have the highest similarity to the ground truth in terms of geometric morphology, gradient variations, and other structural features.

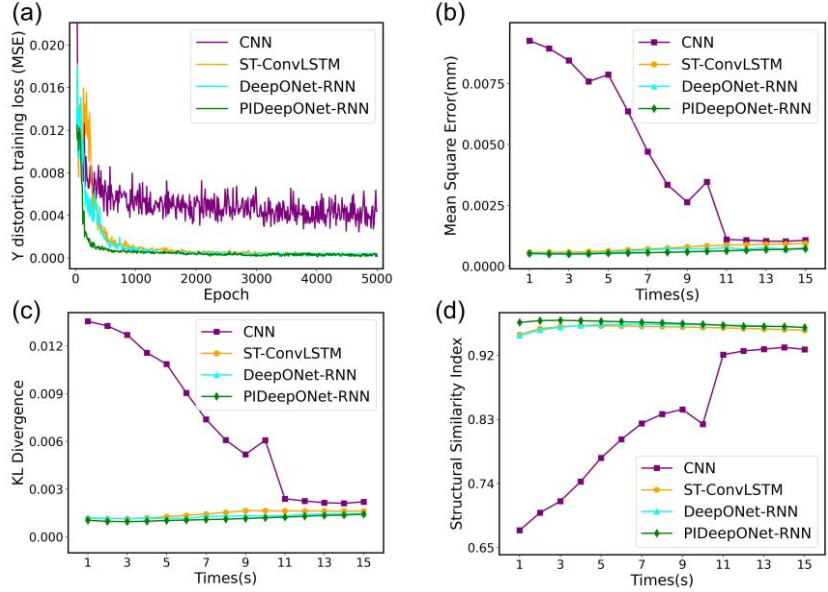

**Fig. 12. Evaluation of surrogate models for y-direction distortion prediction for the future 1-15 s. (a)** Comparison of the training loss. **(b)** Comparison of the MSE. **(c)** Comparison of the KL divergence. **(d)** Comparison of the SSIM.

Figure 12 presents the evaluation results of the four surrogate models for predicting y-direction distortion for the future 1-15 s. In Figure 12(a), the training loss trends are consistent with those for z-direction distortion prediction: the CNN model performs the worst, while PIDeepONet-RNN demonstrates the most stable convergence. This indicates that the influence of model architecture and constraint strategies remains consistent across different distortion directions. Figure 12(b) shows that for y-direction distortion prediction, the MSE of both ST-ConvLSTM and DeepONet-RNN exhibits an increasing trend over time, revealing the error accumulation effect inherent in purely data-driven models when simulating complex distortions. PIDeepONet-RNN alleviates this issue, maintaining a smoother MSE curve at the lowest level. The KL divergence in Figure 12(c) further confirms these findings. KL values of ST-ConvLSTM and DeepONet-RNN increase during the later prediction stages, indicating a deterioration in the physical consistency of their predictions over time. SSIM in Figure 12(d) shows that for y-direction distortion prediction, PIDeepONet-RNN also achieves the highest structural similarity, with its predicted field being geometrically closest to the ground truth. SSIM values of purely data-driven models decrease in the later prediction stages, corresponding to the observed error accumulation phenomenon.

4.4. Comparison of time efficiency in long-horizon prediction

This section analyzes and compares the predicted time efficiency of the four surrogate models with the FEM. The comparison primarily focuses on training and prediction efficiency. Training efficiency is mainly reflected in the model's convergence speed, computational resource requirements, and training time. prediction efficiency focuses on metrics such as model prediction speed, memory footprint, and real-time ability. Table 6 compares the training and predicted time of the four surrogate models with FEM.

**Table 6** Comparison of training and predicted time between four surrogate models and FEM.

|  | FEM | CNN | ST-ConvLSTM | DeepONet-RNN | PIDeepONet-RNN |
|---|---|---|---|---|---|
| Training time | - | 69 min | 67 min | 83 min | 104 min |
| Predicted time | 4 h | 80 ms | 110 ms | 130 ms | 150 ms |

In terms of training efficiency, the training times for the CNN, ST-ConvLSTM, DeepONet-RNN, and PIDeepONet-RNN models are 69 min, 67 min, 83 min, and 104 min, respectively. The shorter training times for CNN and ST-ConvLSTM primarily stem from their simpler network architectures. As a basic feedforward network, CNN has fewer parameters but sacrifices prediction accuracy. Although ST-ConvLSTM introduces a recurrent mechanism, its coupled design maintains fast training speed while limiting its capacity to express complex physical relationships. The training time for DeepONet-RNN increases to 83 minutes, which originates from its decoupled architecture design. The branch and trunk networks need to learn mechanical responses and thermo-dynamic features separately, increasing model complexity and parameter optimization difficulty, but this design enhances the model's prediction accuracy and long-horizon predicting capability. PIDeepONet-RNN has the longest training time because it introduces a physical constraint loss term on the basis of DeepONet-RNN. Each training epoch requires the computation of the physical residual of the heat conduction equation, adding extra computational overhead. However, this mechanism accelerates model convergence through physical regularization and improves prediction accuracy and generalization capability.

In terms of prediction efficiency, all surrogate models possess real-time prediction capability. Compared to the conventional FEM which requires 4 hours to complete a simulation, these trained surrogate models can complete a single long-horizon prediction within ms, meeting the real-time monitoring and control requirements of the WAAM process. More importantly, the surrogate models offer adaptive advantages that are incomparable to the FEM. When facing adjustments in process parameters or changes in geometric sizes, FEM require rebuilding the model, resetting boundary conditions, and performing time-consuming computations, lacking flexibility and adaptive capability. In contrast, the trained surrogate models can directly process new input data and rapidly generate prediction results, providing strong technical support for process optimization and real-time control.

4.5. Discussion

The challenge of long-horizon prediction in thermo-mechanical distortion evolution primarily stems from the cumulative nature of thermo-mechanical coupling effects, the accuracy of ROI predictions, and the path-dependent behavior inherent in WAAM processes. Distortion is not only influenced by instantaneous thermal input but is also profoundly affected by historical thermal cycles and the accumulation of plastic strain. The proposed PIDeepONet-RNN model addresses these challenges by incorporating physical constraints derived from the heat conduction equation, ensuring that the internal thermal dynamics adhere to fundamental physical laws. This enables the model to accurately capture cumulative thermal effects and their mechanical consequences over time.

In contrast, data-driven models such as CNN, ST-ConvLSTM, and even the decoupled DeepONet-RNN rely solely on statistical correlations learned from training data. While they may achieve reasonable short-term accuracy, they lack an inherent understanding of physical causality.

This limitation becomes particularly evident in long-horizon predictions, where errors accumulate and amplify over time, leading to unphysical artifacts and reduced generalization capability. The transition from the 19th to the 20th deposition layer, a phase involving changes in geometric and boundary conditions, further highlighted the differences among the models. CNN model failed completely in this dynamic scenario, while the ST-ConvLSTM and DeepONet-RNN models exhibited blurred predictions near weld boundaries and molten pool area. In contrast, the PIDeepONet-RNN maintained stable and accurate predictions throughout this transition, demonstrating superior robustness and adaptability to evolving process conditions, exhibits more stable ROI prediction results.

The incorporation of physical constraints acts as a powerful regularizer, confining the solution space to physically plausible manifolds and preventing the generation of unphysical predictions. This is especially critical in areas with sparse data or high nonlinearity, such as near heat sources or during layer transitions. The ablation study further confirmed that the decoupled architecture of DeepONet-RNN alone provides improvements over coupled models, but the integration of physics-based learning is essential for achieving both numerical accuracy and physical consistency.

Residual stress, an inevitable consequence of thermo-mechanical effects in metal AM, exhibits a complex and challenging-to-predict distribution[31]. Traditional experimental methods (e.g., X-ray diffraction, neutron diffraction) and numerical simulations exhibit limitations in achieving full-field stress mapping[47]. Recent studies integrating digital image correlation have enabled high-resolution, full-field reconstruction of residual stress in AM parts without simplified assumptions[48]. This study proposes a PIDeepONet-RNN model that incorporates FEM calculations to predict the Von Mises stress during the deposition process. The predicted stress distribution for the future 1 s, compared with the FEM results, is presented in Figure 13. Figure 13(a) compares the FEM results with the stress distribution calculated from the PIDeepONet-RNN predictions, revealing a strong agreement between the numerical predictions and the FEM distribution. Figure 13(b) displays the overall prediction error, which is primarily concentrated within the ROI, with an average error of 2.3%. The max prediction error in the ROI is 9%, as shown in Figure 13(c). This numerical approach facilitates faster and more accurate stress field modeling, aiding researchers in applying the von Mises yield criterion to determine the material's transition from elastic to plastic state. This capability enables early prevention of stress concentration and defect mitigation.

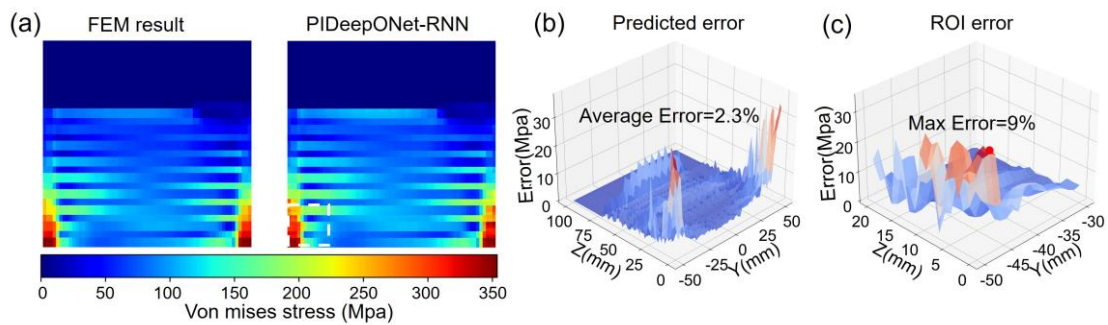

**Fig. 13. Performance of PIDeepONet-RNN model for von mises stress prediction at 1192 s of deposition for the future 1 s. (a)** Comparison of the prediction result with FEM result. **(b)**

Overall prediction error. **(c)** ROI prediction error.

The interpretability of the proposed PIDeepONet-RNN model plays a critical role in promoting its industrial deployment. By embedding the heat conduction equation as a soft physical constraint, the model provides physically meaningful intermediate representations, allowing users to trace the relationship between thermal history, stress evolution, and distortion formation. This feature transforms the model from a purely data-driven predictor into a physics-aware surrogate, offering engineers the ability to analyze the underlying thermo-mechanical mechanisms rather than treating predictions as black-box outputs. Such interpretability is crucial for industrial decision-making, as it facilitates the identification of defect origins, validation of process parameters, and establishment of confidence in automated predictions. From an industrial integration perspective, the proposed surrogate model can be seamlessly embedded into a WAAM DT framework. Its real-time inference speed enables continuous prediction of distortion evolution during layer-by-layer deposition, providing timely feedback for process optimization and defect mitigation. The decoupled architecture also allows flexible interfacing with in-situ sensor data, such as thermal imaging, acoustic emission, or laser scanning, supporting adaptive control strategies and online process corrections. This real-time, physics-informed digital twin integration represents a key step toward intelligent manufacturing, where data-driven learning and physical modeling jointly enable predictive monitoring, reliability assurance, and closed-loop control of metallic AM systems[49].

## 5. Conclusion

This study presents a novel physics-informed neural operator approach for predicting real-time long-horizon distortion evolution in WAAM. The model leverages historical temperature and distortion field data to prediction z and y-direction distortion for the future 15 s. By adopting a decoupled branch-trunk architecture, the model separately captures mechanical response features and thermal dynamics, thereby enhancing both predictive accuracy and physical consistency. The branch network encodes historical distortion fields to extract spatiotemporal physical features, while the trunk network processes historical temperature fields to model thermal evolution. Embedding the heat-conduction equation as a soft constraint on the trunk network ensures that predicted future temperature fields obey physical laws, improving generalization and reducing error accumulation. A comprehensive evaluation of multiple surrogate models, including purely data-driven models (CNN), spatiotemporal-coupled data-driven models (ST-ConvLSTM), neural operator learning models (DeepONet-RNN), and physics-informed neural operator models (PIDeepONet-RNN), demonstrates that the physics-informed neural operator models outperform all others in terms of prediction accuracy, long-horizon stability, and physical plausibility. The main contributions of this study are as follows:

1. PIDeepONet-RNN successfully predicts distortion into the future 15 s for unseen process parameters, achieving a low average MAE. It maintains high accuracy in both spatial directions, with average MAE of 0.0298 mm in z-direction and 0.0157 mm in y-direction. The model also demonstrates robustness even during interlayer transitions.
2. The ConvLSTM units enables the model to learn long-horizon spatiotemporal features and achieve long-horizon distortion prediction for the future 15 s.
3. By integrating the heat conduction equation as a soft constraint into the loss function, the

model ensures that the predicted temperature field of trunk adhere to physical laws. This physics-informed approach suppresses unphysical artifacts and enhances generalization.
4. The decoupled architecture enables more efficient learning. DeepONet-based model achieves convergence with similar accuracy to coupled models using only 60% of the training epochs.
5. The physics-informed neural operator surrogate model provides real-time prediction, offering a rapid and accurate alternative to time-consuming FEM simulations. This efficiency and adaptability make it a promising candidate for real-time monitoring and digital twin applications in AM.

This study opens up several promising research directions for WAAM and broader scientific computing fields, laying the foundation for developing real-time feedforward control and digital twins for the WAAM process. Future applications may include online optimization of process parameters, defect prediction, and closed-loop control. Future research will focus on bridging the "simulation-to-reality" gap by utilizing in-situ monitoring data for model validation, transfer, and iterative improvement to enhance its reliability in the real physical world. From a methodological perspective, future work can explore incorporating deeper physical principles (e.g., thermo-mechanically coupled constitutive relations) into the network and extending it to arbitrary complex 3D geometries, ultimately aiming to achieve a universal, physics-driven scientific ML paradigm.

## References

[1] JIN L, ZHAI X, WANG K, ZHANG K, WU D, NAZIR A, JIANG J, LIAO W-H. Big data, machine learning, and digital twin assisted additive manufacturing: A review [J]. Materials & Design, 2024, 244.

[2] XIONG Y, TANG Y, ZHOU Q, MA Y, ROSEN D W. Intelligent additive manufacturing and design: state of the art and future perspectives [J]. Additive Manufacturing, 2022, 59.

[3] PANT H, ARORA A, GOPAKUMAR G S, CHADHA U, SAEIDI A, PATTERSON A E. Applications of wire arc additive manufacturing (WAAM) for aerospace component manufacturing [J]. The International Journal of Advanced Manufacturing Technology, 2023, 127(11-12): 4995-5011.

[4] SVETLIZKY D, DAS M, ZHENG B, VYATSKIKH A L, BOSE S, BANDYOPADHYAY A, SCHOENUNG J M, LAVERNIA E J, ELIAZ N. Directed energy deposition (DED) additive manufacturing: Physical features, defects, challenges and applications [J]. Materials Today, 2021, 49: 271-95.

[5] KIM S, KIM E-H, LEE W, SIM M, KIM I, NOH J, KIM J-H, LEE S, PARK I, SU P-C, ANDREU A, YOON Y-J. Real-time in-process control methods of process parameters for additive manufacturing [J]. Journal of Manufacturing Systems, 2024, 74: 1067-90.

[6] XIE D, LV F, YANG Y, SHEN L, TIAN Z, SHUAI C, CHEN B, ZHAO J. A Review on Distortion and Residual Stress in Additive Manufacturing [J]. Chinese Journal of Mechanical Engineering: Additive Manufacturing Frontiers, 2022, 1(3).

[7] ZHANG H, LI R, LIU J, WANG K, WEIJIAN Q, SHI L, LEI L, HE W, WU S. State-of-art review on the process-structure-properties-performance linkage in wire arc additive manufacturing [J]. Virtual and Physical Prototyping, 2024, 19(1).

[8] CHEN X, KONG F, FU Y, ZHAO X, LI R, WANG G, ZHANG H. A review on wire-arc

additive manufacturing: typical defects, detection approaches, and multisensor data fusion-based model [J]. The International Journal of Advanced Manufacturing Technology, 2021, 117(3-4): 707-27.

[9]   LI Y, SU C, ZHU J. Comprehensive review of wire arc additive manufacturing: Hardware system, physical process, monitoring, property characterization, application and future prospects [J]. Results in Engineering, 2022, 13.

[10]  WU B, PAN Z, DING D, CUIURI D, LI H, XU J, NORRISH J. A review of the wire arc additive manufacturing of metals: properties, defects and quality improvement [J]. Journal of Manufacturing Processes, 2018, 35: 127-39.

[11]  CHANDRA S, RADHAKRISHNAN J, HUANG S, WEI S, RAMAMURTY U. Solidification in metal additive manufacturing: challenges, solutions, and opportunities [J]. Progress in Materials Science, 2025, 148.

[12]  MIRAZIMZADEH S E, MOHAJERNIA B, PAZIREH S, URBANIC J, JIANU O. Investigation of residual stresses of multi-layer multi-track components built by directed energy deposition: experimental, numerical, and time-series machine-learning studies [J]. The International Journal of Advanced Manufacturing Technology, 2023, 130(1-2): 329-51.

[13]  LI R, JU G, ZHAO X, ZHANG Y, LI Y, HU G, YAN M, WU Y, LIN D. Simulation of residual stress and distortion evolution in dual-robot collaborative wire-arc additive manufactured Al-Cu alloys [J]. Virtual and Physical Prototyping, 2024, 19(1).

[14]  CADIOU S, COURTOIS M, CARIN M, BERCKMANS W, LE MASSON P. 3D heat transfer, fluid flow and electromagnetic model for cold metal transfer wire arc additive manufacturing (Cmt-Waam) [J]. Additive Manufacturing, 2020, 36.

[15]  LI R, XIONG J, LEI Y. Investigation on thermal stress evolution induced by wire and arc additive manufacturing for circular thin-walled parts [J]. Journal of Manufacturing Processes, 2019, 40: 59-67.

[16]  CHEN X, WANG C, DING J, BRIDGEMAN P, WILLIAMS S. A three-dimensional wire-feeding model for heat and metal transfer, fluid flow, and bead shape in wire plasma arc additive manufacturing [J]. Journal of Manufacturing Processes, 2022, 83: 300-12.

[17]  OU W, WEI Y, LIU R, ZHAO W, CAI J. Determination of the control points for circle and triangle route in wire arc additive manufacturing (WAAM) [J]. Journal of Manufacturing Processes, 2020, 53: 84-98.

[18]  BAYAT M, DONG W, THORBORG J, TO A C, HATTEL J H. A review of multi-scale and multi-physics simulations of metal additive manufacturing processes with focus on modeling strategies [J]. Additive Manufacturing, 2021, 47.

[19]  CHEN Z, YUAN L, PAN Z, ZHU H, MA N, DING D, LI H. A comprehensive review and future perspectives of simulation approaches in wire arc additive manufacturing (WAAM) [J]. International Journal of Extreme Manufacturing, 2025, 7(2).

[20]  HE F, YUAN L, MU H, ROS M, DING D, PAN Z, LI H. Research and application of artificial intelligence techniques for wire arc additive manufacturing: a state-of-the-art review [J]. Robotics and Computer-Integrated Manufacturing, 2023, 82.

[21]  PARSAZADEH M, SHARMA S, DAHOTRE N. Towards the next generation of machine learning models in additive manufacturing: A review of process dependent material evolution [J]. Progress in Materials Science, 2023, 135.

[22]  QIN J, HU F, LIU Y, WITHERELL P, WANG C C L, ROSEN D W, SIMPSON T W, LU Y,

TANG Q. Research and application of machine learning for additive manufacturing [J]. Additive Manufacturing, 2022, 52.

[23] ZHANG R, STRICKLAND J, HOU X, YANG F, LI X, DE OLIVEIRA J A, LI J, ZHANG S. Rapid residual stress simulation and distortion mitigation in laser additive manufacturing through machine learning [J]. Additive Manufacturing, 2025, 102.

[24] HAJIALIZADEH F, INCE A. Integration of artificial neural network with finite element analysis for residual stress prediction of direct metal deposition process [J]. Materials Today Communications, 2021, 27.

[25] HAJIALIZADEH F, INCE A. Residual stress computation in direct metal deposition using integrated artificial neural networks and finite element analysis [J]. Materials Today Communications, 2024, 38.

[26] WU Q, MUKHERJEE T, DE A, DEBROY T. Residual stresses in wire-arc additive manufacturing – Hierarchy of influential variables [J]. Additive Manufacturing, 2020, 35.

[27] XIE X, BENNETT J, SAHA S, LU Y, CAO J, LIU W K, GAN Z. Mechanistic data-driven prediction of as-built mechanical properties in metal additive manufacturing [J]. npj Computational Materials, 2021, 7(1).

[28] ZHOU Z, SHEN H, LIN J, LIU B, SHENG X. Continuous tool-path planning for optimizing thermo-mechanical properties in wire-arc additive manufacturing: An evolutional method [J]. Journal of Manufacturing Processes, 2022, 83: 354-73.

[29] ZHOU Z, SHEN H, LIU B, DU W, JIN J, LIN J. Residual thermal stress prediction for continuous tool-paths in wire-arc additive manufacturing: a three-level data-driven method [J]. Virtual and Physical Prototyping, 2021, 17(1): 105-24.

[30] MU H, HE F, YUAN L, HATAMIAN H, COMMINS P, PAN Z. Online distortion simulation using generative machine learning models: A step toward digital twin of metallic additive manufacturing [J]. Journal of Industrial Information Integration, 2024, 38.

[31] LI X, ZHAO Z, CHENG X, CHEN H, XIONG J. Residual stress and distortion in arc-directed energy deposition: Formation mechanisms, analysis techniques, and mitigation strategies [J]. Journal of Manufacturing Processes, 2025, 141: 17-35.

[32] HAMRANI A, AGARWAL A, ALLOUHI A, MCDANIEL D. Applying machine learning to wire arc additive manufacturing: a systematic data-driven literature review [J]. Journal of Intelligent Manufacturing, 2023, 35(6): 2407-39.

[33] LI Z, ZHENG H, KOVACHKI N, JIN D, CHEN H, LIU B, AZIZZADENESHELI K, ANANDKUMAR A. Physics-Informed Neural Operator for Learning Partial Differential Equations [J]. ACM / IMS Journal of Data Science, 2024, 1(3): 1-27.

[34] KARNIADAKIS G E, KEVREKIDIS I G, LU L, PERDIKARIS P, WANG S, YANG L. Physics-informed machine learning [J]. Nature Reviews Physics, 2021, 3(6): 422-40.

[35] SHENGHAN G, MOHIT A, CLAYTON C, QI T, X. G R, WEIHONG G, Y.B. G. Machine learning for metal additive manufacturing: Towards a physics-informed data-driven paradigm [J]. Journal of Manufacturing Systems, 2022, 62: 145-63.

[36] SHARMA R, GUO Y B. Thermo-mechanical physics-informed deep learning for prediction of thermal stress evolution in laser metal deposition [J]. Engineering Applications of Artificial Intelligence, 2025, 157.

[37] LU L, JIN P, PANG G, ZHANG Z, KARNIADAKIS G E. Learning nonlinear operators via DeepONet based on the universal approximation theorem of operators [J]. Nature Machine

Intelligence, 2021, 3(3): 218-29.

[38] JIAO W, ZHAO D, YANG S, XU X, ZHANG X, LI L, CHEN H. Real-time prediction of temperature field during welding by data-mechanism driving [J]. Journal of Manufacturing Processes, 2025, 133: 260-70.

[39] KUSHWAHA S, PARK J, KORIC S, HE J, JASIUK I, ABUEIDDA D. Advanced deep operator networks to predict multiphysics solution fields in materials processing and additive manufacturing [J]. Additive Manufacturing, 2024, 88.

[40] WANG S, WANG H, PERDIKARIS P. Learning the solution operator of parametric partial differential equations with physics-informed DeepONets [J]. Science Advances, 2021, 7(40).

[41] GOLDAK J, CHAKRAVARTI A, BIBBY M. A new finite element model for welding heat sources [J]. Metallurgical Transactions B, 1984, 15(2): 299-305.

[42] TIAN M, MU H, LIU T, LI M, DING D, ZHAO J. Physics-informed machine learning-based real-time long-horizon temperature fields prediction in metallic additive manufacturing [J]. Communications Engineering, 2025, 4(1).

[43] GAO H, LI H, SHAO D, FANG N, MIAO Y, PAN Z, LI H, ZHANG B, PENG Z, WU B. Towards quality controllable strategies in wire-arc directed energy deposition [J]. International Journal of Extreme Manufacturing, 2025, 7(4).

[44] ZHAO T, YAN Z, ZHANG B, ZHANG P, PAN R, YUAN T, XIAO J, JIANG F, WEI H, LIN S, CHEN S. A comprehensive review of process planning and trajectory optimization in arc-based directed energy deposition [J]. Journal of Manufacturing Processes, 2024, 119: 235-54.

[45] SHI X, CHEN Z, WANG H, YEUNG D-Y, WONG W-K, WOO W-C. Convolutional LSTM Network: A Machine Learning Approach for Precipitation Nowcasting [J]. Computer Science, 2015.

[46] WANG Y, WU H, ZHANG J, GAO Z, WANG J, YU P S, LONG M. PredRNN: A Recurrent Neural Network for Spatiotemporal Predictive Learning [J]. IEEE Transactions on Pattern Analysis and Machine Intelligence, 2023, 45(2): 2208-25.

[47] BARTLETT J L, CROOM B P, BURDICK J, HENKEL D, LI X. Revealing mechanisms of residual stress development in additive manufacturing via digital image correlation [J]. Additive Manufacturing, 2018, 22: 1-12.

[48] UZUN F, BASOALTO H, LIOGAS K, CHEN J, DOLBNYA I P, WANG Z I, KORSUNSKY A M. Voxel-based full-field eigenstrain reconstruction of residual stresses in additive manufacturing parts using height digital image correlation [J]. Additive Manufacturing, 2023, 77.

[49] MU H, HE F, YUAN L, COMMINS P, WANG H, PAN Z. Toward a smart wire arc additive manufacturing system: A review on current developments and a framework of digital twin [J]. Journal of Manufacturing Systems, 2023, 67: 174-89.